\documentclass[10pt,twocolumn,letterpaper]{article}

\usepackage[pagenumbers]{cvpr} 
\usepackage{booktabs}
\usepackage{multirow}
\usepackage{cellspace}
\usepackage{makecell}
\usepackage{lipsum}
\usepackage{bm}
\usepackage{relsize} 
\usepackage{booktabs, multirow, array, makecell}
\usepackage{soul}
\usepackage{graphicx} 
\usepackage{mathtools} 
\usepackage{colortbl}
\usepackage{xcolor} 
\usepackage{subcaption}
\usepackage{wrapfig}
\usepackage{natbib}
\usepackage{float}
\usepackage{amssymb}
\usepackage{pifont}
\newcommand{\cmark}{\ding{51}}%
\definecolor{cvprblue}{rgb}{0.21,0.49,0.74}
\usepackage[pagebackref,breaklinks,colorlinks,allcolors=cvprblue]{hyperref}

\def\paperID{} 
\def\confName{CVPR}
\def\confYear{2025}

\title{Occlusion-aware Text-Image-Point Cloud Pretraining \\for Open-World 3D Object Recognition}

\author{Khanh Nguyen, Ghulam Mubashar Hassan \& Ajmal Mian \\
The University of Western Australia \\
\texttt{duykhanh.nguyen@research.uwa.edu.au} \\ 
\texttt{\{ghulam.hassan,ajmal.mian\}@uwa.edu.au} \\
}

\begin{document}
\maketitle
\begin{abstract}
\label{sec:abstract}
Recent open-world representation learning approaches have leveraged CLIP to enable zero-shot 3D object recognition. However, performance on real point clouds with occlusions still falls short due to unrealistic pretraining settings. Additionally, these methods incur high inference costs because they rely on Transformer's attention modules.
In this paper, we make two contributions to address these limitations. First, we propose occlusion-aware text-image-point cloud pretraining to reduce the training-testing domain gap. From 52K synthetic 3D objects, our framework generates nearly 630K partial point clouds for pretraining, consistently improving real-world recognition performances of existing popular 3D networks.
Second, to reduce computational requirements, we introduce DuoMamba, a two-stream linear state space model tailored for point clouds. By integrating two space-filling curves with 1D convolutions, DuoMamba effectively models spatial dependencies between point tokens, offering a powerful alternative to Transformer.
When pretrained with our framework, DuoMamba surpasses current state-of-the-art methods while reducing latency and FLOPs, highlighting the potential of our approach for real-world applications. Our code and data are available at 
\href{https://ndkhanh360.github.io/project-occtip}{\texttt{ndkhanh360.github.io/project-occtip}}.
\end{abstract}
\vspace{-5mm}    
\section{Introduction}
\label{sec:intro}
\begin{figure*}[!t]
    \centering
    \includegraphics[width=\linewidth]{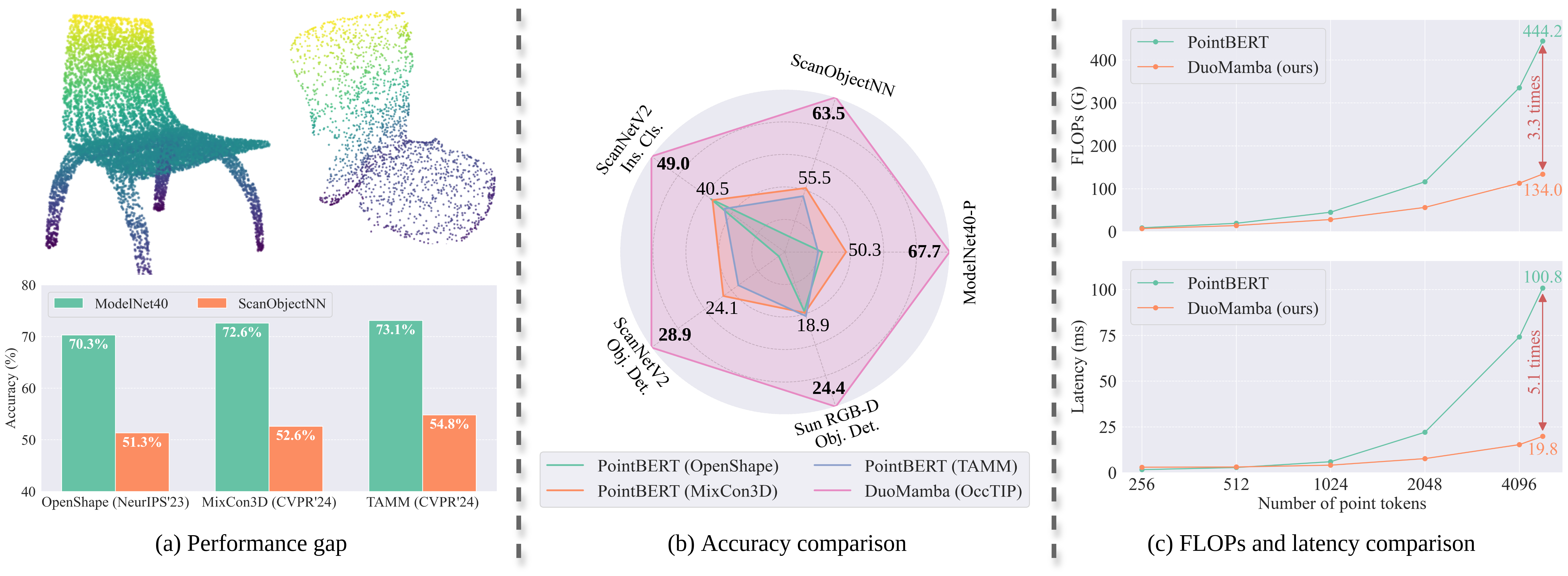}
    \vspace{-9mm}
    \caption{Comparison to existing methods. (a) State-of-the-art approaches pretrain 3D encoders on complete point clouds, which differ significantly from occluded ones in practical scenarios (top). This leads to a substantial gap in zero-shot performance between ModelNet40 \cite{modelnet40} benchmark with full point clouds and ScanObjectNN \cite{scanobjectnn} with real-world data (bottom). (b) The proposed framework OccTIP pretrains 3D models on partial point clouds to better simulate practical conditions, leading to significant improvements on various recognition tasks, especially when combined with our DuoMamba architecture. (c) Compared to the popular PointBERT \cite{pointbert}, DuoMamba has significantly lower FLOPs (top) and latency (bottom) during inference, making it better suited for real-world applications.}    
    \label{fig:fig1_motivation}
    \vspace{-6mm}
\end{figure*}
3D understanding plays a vital role in robotics \cite{affordance}, virtual reality \cite{semantic_parsing}, and autonomous driving \cite{vehicle_detection}, enabled by deep-learning models that perform recognition tasks such as 3D object classification \cite{pointnet}, object detection \cite{3detr,votenet}, and semantic segmentation \cite{octree_cnn,swan}. However, existing 3D networks \cite{pointnet,pointnet++,3detr,dgcnn,point_trans_v3} are trained using closed-set annotation, constraining them to recognize only pre-defined categories and struggle with `unseen' ones. Inspired by CLIP \cite{2dclip}, recent open-world studies \cite{clip2point,pointclip,pointclipv2,openshape,uni3d,ulip} have extended the aligned image-text latent space to include 3D object representations, allowing generalization beyond `seen' categories and enabling zero-shot 3D recognition.

Existing works in this line of research take 3D-image-text triplets as input and align the three embedding spaces using cross-modal contrastive learning. These methods represent 3D shapes either as depth maps \cite{opendlign,clip2point,pointclip,pointclipv2} or raw point clouds \cite{openshape,tamm,ulip,ulip2,uni3d,mixcon3d}. Depth-based approaches must first convert point clouds into 2D depth maps and use pretrained image encoders, such as Vision Transformer (ViT) \cite{vit}, for 3D feature extractions. However, their performance typically suffers from information loss during the projection and the domain gap caused by differences between RGB and depth images.
On the other hand, point-based methods \cite{ulip,openshape,ulip2,uni3d,mixcon3d,tamm}  can directly exploit all intrinsic geometry in the point clouds. An example is the recent work CLIP$^2$ \cite{clip2}, which pretrains a 3D encoder using real-scanned objects extracted from scene-level point clouds. For contrastive learning, it pairs these with cropped images and simple category-based prompts as text descriptions. 
However, limited caption diversity and poor cropped image quality (due to occlusion, lighting, etc.) hinder CLIP knowledge transfer, leading to suboptimal performance.

Other works \cite{ulip,ulip2,openshape,uni3d,mixcon3d,tamm} instead leverage synthetic 3D models\footnote{In this paper, we use \textit{3D models} to refer to 3D CAD models or 3D meshes instead of 3D deep learning models.} to construct pretraining triplets. These methods uniformly sample points from the mesh surface to create full point clouds\footnote{A \textit{full} (\textit{complete}) point cloud provides 360-degree coverage of an object, while a \textit{partial} (\textit{occluded}) one is captured from a single viewpoint.}. They also render RGB images from preset camera positions and generate diverse captions from multiple sources, allowing for control over the quality of images and texts. As a result, these methods demonstrate promising zero-shot performance on complete point cloud benchmarks such as ModelNet40 \cite{modelnet40}. However, their performance degrades significantly on real-scanned data, leading to unsatisfactory results in practical scenarios.  As shown in Figure \ref{fig:fig1_motivation}a, there is a 20\% accuracy drop from the synthetic ModelNet40 \cite{modelnet40} to the real ScanObjectNN \cite{scanobjectnn}, caused by the large domain gap between complete point clouds in pretraining and occluded ones encountered in real-world conditions.
To address this data discrepancy, we introduce an occlusion-aware pretraining framework that leverages synthetic 3D meshes to create partial point clouds. We simulate real-world scenarios by putting a virtual camera around an object and only sample points visible from the camera position. From 52K ShapeNetCore \cite{shapenet} 3D models, our framework generates nearly 630K occluded point clouds for pretraining and enhances the zero-shot accuracy of SparseConv \cite{sparseconv} and PointBERT \cite{pointbert} by 3.8\% and 5.1\% on ScanObjectNN \cite{scanobjectnn}. Despite using only synthetic objects, our framework consistently improves recognition performance on various real-world tasks and even outperforms methods that use real-scanned data.

\vspace{-1mm}
Moreover, existing multi-modal pretraining approaches \cite{mixcon3d,uni3d,openshape,ulip,ulip2} heavily rely on Transformer-based 3D encoders due to their strong learning capacity. However, these pretrained models have high inference costs because of the attention's quadratic complexity. This poses significant challenges when we want to increase the number of point tokens in the input or use the pretrained encoder as a classification head in a 3D object detector. Inspired by Mamba \cite{mamba}, we introduce an efficient architecture named DuoMamba as an alternative to Transformer-based models. At the core of our network is the two-stream DuoMamba block, developed using linear-time S6 modules from Mamba \cite{mamba}. Each stream processes point tokens in the order from a space-filling curve, either Hilbert \cite{hilbert_curve} or its transposed variant Trans-Hilbert. Intuitively, these turn an unordered point cloud into a geometrically structured sequence where close points in 3D space stay adjacent in the sequence, facilitating S6 to capture meaningful geometric relationships. We also replace causal 1D convolutions commonly used in Mamba models \cite{mamba,vmamba,mamba3d,pointmamba} with standard 1D convolutions to allow point tokens to aggregate information of their neighbors in both directions, enriching their spatial context. 
Compared to the popular Transformer-based PointBERT \cite{pointbert}, our model achieves higher performance across several benchmarks (Figure \ref{fig:fig1_motivation}b) while significantly reducing FLOPs and latency (Figure \ref{fig:fig1_motivation}c). It also exhibits a better performance-computation balance than existing Mamba-based point cloud networks \cite{mamba3d,pointmamba}. In summary, our main contributions are:
\begin{itemize}
\item We propose an occlusion-aware pretraining framework for open-world 3D recognition. By generating partial point clouds from synthetic 3D models, our approach simulates real-world conditions and removes the need for real-scanned data in pretraining.
\item We demonstrate, through extensive experiments, that our framework consistently improves the performance of two popular networks: PointBERT \cite{pointbert} and SparseConv \cite{sparseconv}.
\item We introduce DuoMamba, a two-stream linear-time architecture integrated with space-filling curves and 1D convolutions for efficient point cloud learning. Our network achieves higher accuracy than Transformer-based methods, with reduced computation and lower latency.

\end{itemize}
\begin{figure*}[!t]
    \centering
    \includegraphics[width=\linewidth]{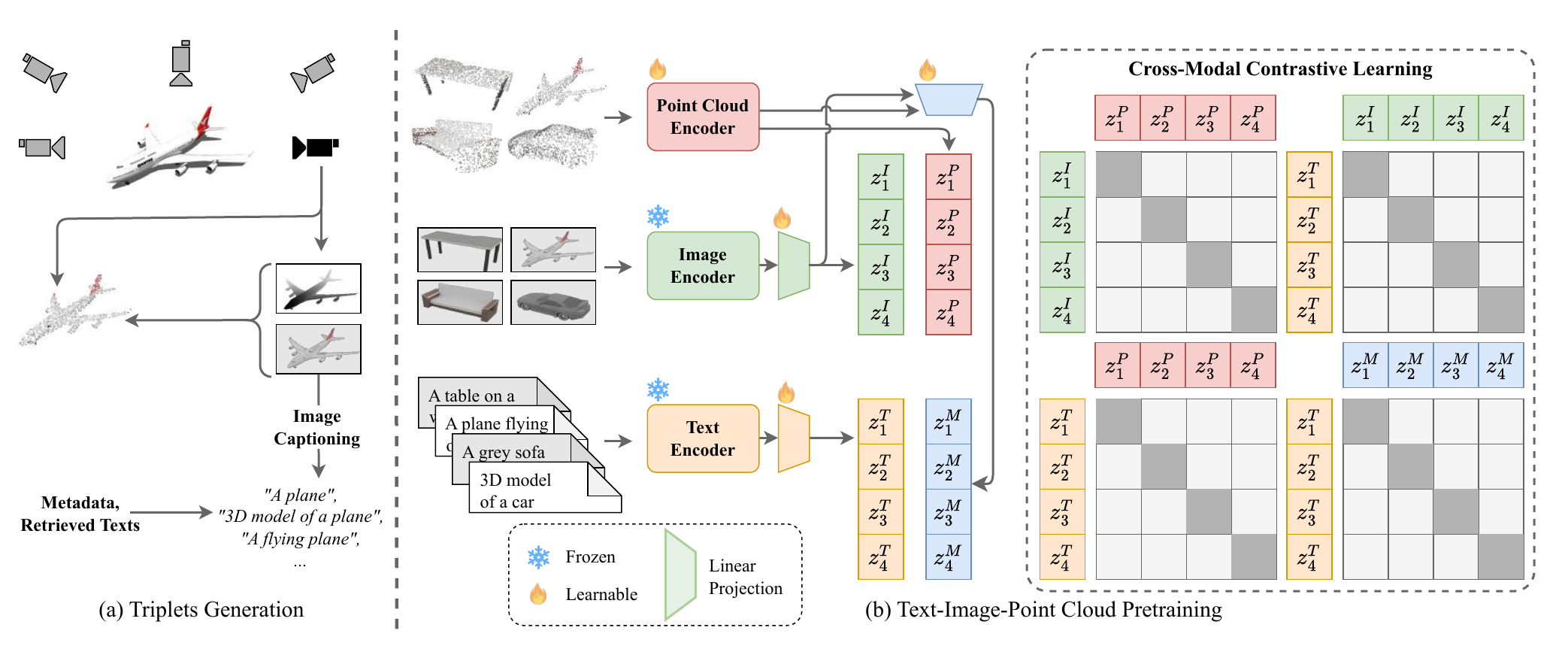}
    \vspace{-10mm}
    \caption{Overview of our OccTIP pretraining framework. (a) Given a 3D object, we generate RGB and depth images from preset camera positions, which are used to construct partial point clouds. Texts are generated from dataset metadata, image captioning models \cite{blip}, and retrieved descriptions of similar photos from LION-5B \cite{laion_5b}. (b) During pretraining, we extract multi-modal features using a learnable point cloud network and frozen CLIP \cite{2dclip} encoders, then align them through contrastive learning.}
    \label{fig:fig2_pretraining_framework}
    \vspace{-5mm}    
\end{figure*}
\vspace{-2mm}
\section{Related Work}
\label{sec:related_work}
\noindent\textbf{CLIP for 3D Representation Learning.}
Vision-Language Models (VLMs) such as CLIP \cite{2dclip} and ALIGN \cite{align} have demonstrated impressive zero-shot capabilities through contrastive learning on large image-text corpora. These models effectively map the two modalities into a shared latent space with rich semantic and visual concepts, forming a foundation for various 2D applications \cite{detic,dalle,sam,imagegen}. Recently, several studies have leveraged CLIP for 3D representation learning, showing promising results in object-level zero-shot 3D recognition \cite{ulip,openshape,pointclip,clip2point,clip2,mixcon3d,tamm,opendlign}.

Among them, several works \cite{pointclip,pointclipv2,clip2point,opendlign} project point clouds into depth maps and rely on fine-tuning CLIP image encoders for zero-shot classification. However, they often experience information loss during 3D-to-depth projections, which significantly impacts their performance. In contrast, other methods \cite{clip2,ulip,ulip2,openshape,uni3d,tamm,mixcon3d} train specialized point cloud encoders to distill CLIP knowledge, extending the image-text co-embedding space to encompass 3D representations. These approaches form text-image-point cloud triplets and utilize contrastive learning to align the latent spaces of the three modalities. For instance, CLIP$^2$ \cite{clip2} uses object point clouds and images from real scenes to generate pretraining triplets. However, the quality of the cropped images can vary due to lighting conditions, object size, and occlusion. Also, object descriptions are created from simple prompts, leading to suboptimal transfer of CLIP knowledge and unsatisfactory performance. 
Other works \cite{ulip,ulip2,openshape,uni3d,tamm,mixcon3d} use synthetic 3D models to render RGB images and leverage metadata, image captioning models \cite{blip}, and retrieved texts for diverse descriptions. However, they typically pretrain 3D encoders on complete point clouds, which greatly differ from the real ones encountered in practical conditions due to occlusion and viewpoint limitations. To address this, we propose a framework that uses synthetic 3D models to generate occluded point clouds for pretraining, reducing data discrepancies while maintaining high image quality and caption diversity for effective transfer of CLIP's knowledge.
\begin{figure*}[!t]
    \centering
    \includegraphics[width=\linewidth]{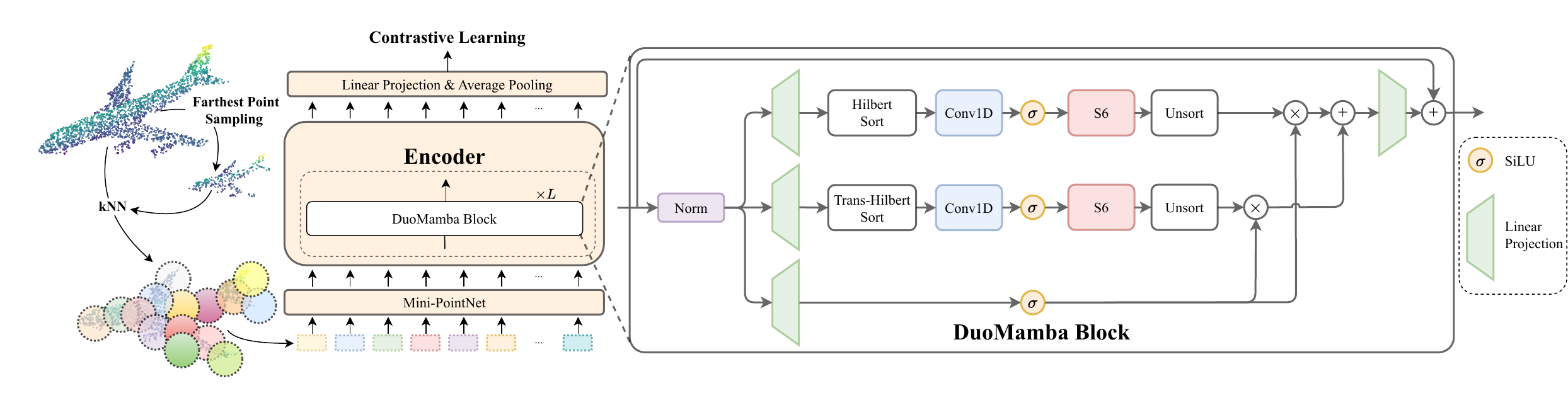}
    \vspace{-8mm}
    \caption{Overview of the proposed architecture and detailed design of our DuoMamba block. We integrate two Hilbert curves \cite{hilbert_curve} and standard 1D convolutions with linear-time S6 \cite{mamba} modules to efficiently model geometric dependencies and enrich spatial context. }
    \label{fig:fig3_proposed_model}
    \vspace{-5mm}    
\end{figure*}

\vspace{1mm} \noindent\textbf{Deep Learning-Based Point Cloud Encoders.}
Leveraging deep learning, the pioneering PointNet \cite{pointnet} directly processes point clouds using multi-layer perceptrons applied on each point independently. Subsequent methods \cite{pointnet++,pointnext,dgcnn} introduce hierarchical structures to model local neighborhoods and geometric relationships, addressing PointNet's limitations. Alternatively, convolution-based approaches \cite{voxnet,sparseconvnet} convert point clouds into 3D voxel grids, utilizing established 3D convolutions for feature learning. SparseConv \cite{sparseconvnet} reduces the high memory requirements of 3D convolutions through sparse convolution, enhancing the voxel-based method's applicability. Since the introduction of self-attention in Transformers \cite{transformer}, most state-of-the-art encoders \cite{pointbert,pointmae,point_trans_v2,point_trans_v3} are based on this architecture, with PointBERT \cite{pointbert} being a representative for object-level point cloud pretraining \cite{pointbert,openshape,ulip,ulip2,tamm}. However, the attention mechanism's quadratic complexity results in high computational costs as the input length increases.

To overcome this, Mamba3D \cite{mamba3d} and PointMamba \cite{pointmamba} were developed using the linear-time S6 from Mamba \cite{mamba} as alternatives to attention layers. However, these networks overlook key characteristics of point clouds. Specifically, Mamba3D \cite{mamba3d} applies S6 to point tokens in random order due to the unstructured nature of point clouds, which is not optimal since S6 was designed for sequence data with meaningful order, such as natural language and audio. PointMamba \cite{pointmamba} improves on this by sorting points using Hilbert and Trans-Hilbert curves \cite{hilbert_curve}, ensuring that spatially close points remain adjacent in the sequence. However, it simply concatenates the two resulting orders as input for S6, doubling the sequence length and computations. Moreover, both methods employ causal 1D convolution, which is beneficial for causal data like audio but suboptimal for spatial data.
Therefore, we propose a new Mamba-based architecture that integrates point cloud properties into its design and leverages multi-modal pretraining to enhance model knowledge and extend its applicability.
\section{Preliminaries}
\label{sec:preliminaries}
\vspace{1mm} \noindent\textbf{State Space Model}  
represents a continuous system that maps an input $x_{t}$ to an output $y_{t}$ via an implicit latent state $h_{t} \in \mathbb{R}^N$. S4 \cite{s4} introduces a discretized version for sequence-to-sequence transformation, defined as: 
\begin{equation} \label{eq:ssm}
h_{t} = \overline{\boldsymbol{A}} h_{t-1} + \overline{\boldsymbol{B}} x_{t},\qquad y_{t} = \boldsymbol{C}h_{t},
\end{equation}
where $\overline{\boldsymbol{A}}$ and $\overline{\boldsymbol{B}}$ are derived from the model parameters ($\boldsymbol{A}, \boldsymbol{B}, \boldsymbol{C}, \boldsymbol{\Delta}$) using zero-order hold discretization. 
As the update matrices $\overline{\boldsymbol{A}}, \overline{\boldsymbol{B}}, \boldsymbol{C}$ are shared across time steps, S4 achieves linear-time computation through a convolution kernel, though its capacity to capture dynamic input sequences is limited. To improve context awareness, the Selective SSM (S6) introduced in Mamba \cite{mamba} makes $\boldsymbol{B}, \boldsymbol{C}, \boldsymbol{\Delta}$ dependent on the input. To maintain near-linear time complexity, Mamba \cite{mamba} employs a hardware-aware implementation for S6, which we follow to ensure computational efficiency. For further details, please refer to \cite{mamba}.

\vspace{1mm} \noindent\textbf{Space-Filling Curves} 
pass through every point in a high-dimensional space, preserving spatial proximity of the original structure. For point clouds, they can be defined as a bijective function \( \Phi: \mathbb{Z}^3 \to \mathbb{Z} \), mapping each \((x, y, z)\) coordinate to a position in a 1D sequence. Our DuoMamba leverages the Hilbert space-filling curve \cite{hilbert_curve} and its transposed variant (Trans-Hilbert) for their strong locality-preserving properties, ensuring that points close in 3D space remain adjacent in the sequence. This is especially valuable for point cloud processing, where points are inherently unordered, making it challenging for sequence models like S6 to capture geometric relationships. By establishing a meaningful order with Hilbert curves, we enable S6 to model spatial dependencies in point clouds more effectively.

\vspace{1mm} \noindent\textbf{Cross-Modal Contrastive Learning.}
\label{subsec:contrastive_learning}
CLIP \cite{2dclip} is a pioneering approach that employs cross-modal contrastive learning to align embeddings of the same concept across two modalities (\eg, a caption \textit{``this is a dog"} and an image of a dog) by pulling their representations closer in a shared-embedding space while pushing apart those of different concepts. Formally, for a batch of $B$ paired features from two modalities $M_1$ and $M_2$, represented as $\{(z_{i}^{M_1}, z_{i}^{M_2})\}_{i=1}^{B}$, the training objective is to minimize the contrastive loss $\mathcal{L}^{M_1 \leftrightarrow M_2}$, defined as: 
\begin{equation}
\label{eq:contrastive_loss}
\begin{aligned}
\mathcal{L}^{M_1 \leftrightarrow M_2} = -\frac{1}{2}(l^{M_1 \rightarrow M_2} + l^{M_2 \rightarrow M_1}),
\end{aligned}
\end{equation}
with $l^{M_1 \rightarrow M_2}$ and $l^{M_2 \rightarrow M_1}$ calculated as follows:
\begin{equation}
\begin{aligned}
    l^{M_1 \rightarrow M_2} &= \sum_{i=1}^{B} \log \frac{\exp(z_i^{M_1} \cdot z_i^{M_2} / \tau)}{\sum_{j=1}^{B} \exp(z_i^{M_1} \cdot z_j^{M_2} / \tau)}, \\ 
    l^{M_2 \rightarrow M_1} &= \sum_{i=1}^{B} \log \frac{\exp(z_i^{M_2} \cdot z_i^{M_1} / \tau)}{\sum_{j=1}^{B} \exp(z_i^{M_2} \cdot z_j^{M_1} / \tau)},    
\end{aligned}
\label{eq:component_loss}
\end{equation}
where \( \tau \) is a temperature parameter that controls the sharpness of the Softmax distributions during training.
\section{Pretraining Framework}
\label{sec:occtip}
\begin{figure*}[t]
    \centering
    \includegraphics[width=\linewidth]{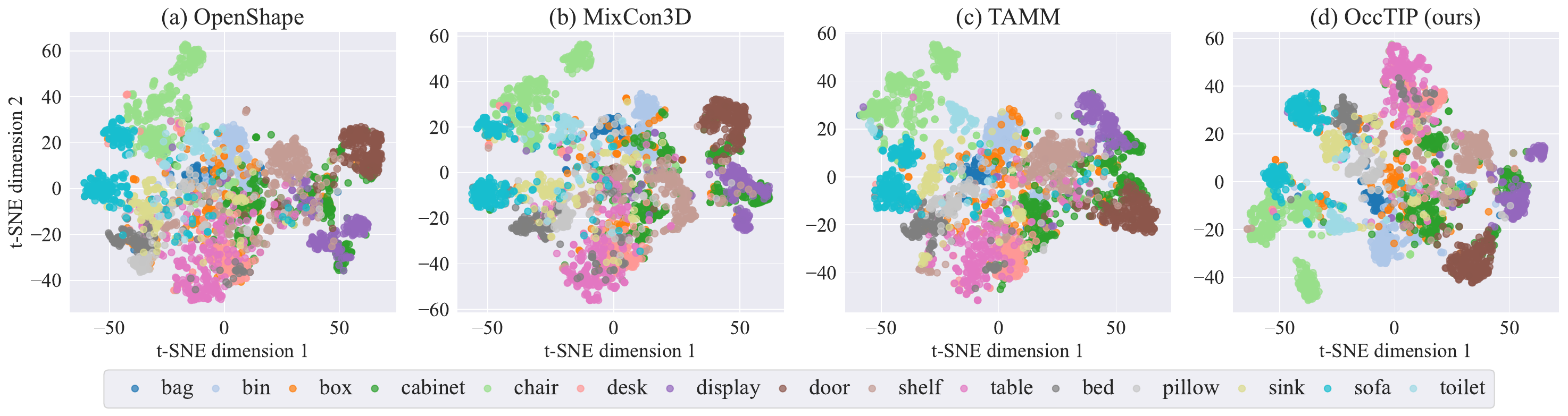}
    \vspace{-7mm}
    \caption{t-SNE visualization of ScanObjectNN \cite{scanobjectnn} features extracted by different pretraining methods. Compared to other approaches based on complete point clouds, our method OccTIP achieves clearer class separation and significantly reduces overlap between classes.}
    \label{fig:tsne}
    \vspace{-5mm}
\end{figure*}
\vspace{1mm} \noindent\textbf{Triplets Generation.}
Given a 3D model, we first center and normalize it to lie within a unit sphere. Following OpenShape \cite{openshape}, we select 12 camera positions uniformly distributed around the object. They lie on a sphere of radius 2, with four viewpoints above the mesh ($z > 0$), four at the same level ($z = 0$), and four below ($z < 0$) to cover all angles. From each position, we render an RGB image and a depth map using BlenderProc \cite{blenderproc}. We set up the scene with area light and use Blender's ray-tracing render engine `CYCLES' for more realistic output. We then construct a partial point cloud based on the camera position, color information from the RGB image, and geometric information in the depth map. On a single RTX 4090 GPU, it takes around three days to process 52K ShapeNetCore \cite{shapenet} 3D models.

For language modality, we use the captions provided by OpenShape \cite{openshape}, which come from three sources: (1) metadata of the dataset, (2) captions generated by BLIP \cite{blip} and Azure cognitive services, and (3) retrieved captions of visually similar images from LAION-5B dataset \cite{laion_5b}. An illustration of the generation process is shown in Figure \ref{fig:fig2_pretraining_framework}a.

\vspace{1mm} 
\noindent\textbf{Training Pipeline.}
For the $i$-th object, we randomly select the partial point cloud and RGB image corresponding to a viewpoint, along with a text from its available captions, to form a triplet $x_i = (T_i, I_i, P_i)$. Input at each iteration is a batch of $B$ triplets, represented as $\{(T_i, I_i, P_i)\}_{i=1}^B$.

The framework trains a point cloud network \( f^P \) to learn 3D representations that are aligned with the embedding spaces of language and images. To achieve this, we leverage pretrained text and image encoders from CLIP \cite{openclip}, denoted as \( f^T \) and \( f^I \), to generate prior features that serve as anchors in the new co-embedding space. Since CLIP \cite{2dclip} was trained on a large image-text corpus and provides a well-aligned latent space, we freeze the primary CLIP \cite{2dclip} encoders during training. To enable flexible alignment of this shared latent space with the additional 3D modality, we introduce learnable projection heads \( h^I \) and \( h^T \) for image and text inputs. They will be updated jointly with the point cloud model during pretraining. Given an input batch, we extract features for each modality as follows:
\begin{equation}
    z^T_i = h^T \left( f^T(T_i) \right), \;
    z^I_i = h^I \left( f^I(I_i) \right), \;
    z^P_i = f^P(P_i).
\end{equation}

Inspired by MixCon3D \cite{mixcon3d}, we introduce additional mixed representations to enhance contrastive learning constraints. Specifically, we compute a combined embedding from the point cloud and image features as:
\begin{equation}
    z^M_i = h^M \left( \mathrm{Concat}(z^P_i, z^I_i) \right),
\end{equation}
where \( h^M \) is a learnable linear projection mapping the concatenated features to the shared latent space.

Finally, we employ cross-modal contrastive learning to `pull' multi-modal features together. The training objective is to minimize the following total loss:
\begin{equation}
    \mathcal{L} = \mathcal{L}^{P \leftrightarrow I} + \mathcal{L}^{P \leftrightarrow T} + \mathcal{L}^{I \leftrightarrow T} + \mathcal{L}^{M \leftrightarrow T},
\end{equation}
where \( T \), \( I \), \( P \), and \( M \) denote text, image, point cloud, and mixed modalities, with each contrastive loss defined in Equation \ref{eq:contrastive_loss}. Our training pipeline is illustrated in Figure \ref{fig:fig2_pretraining_framework}b.

\section{DuoMamba}
\label{sec:duomamba}
\vspace{1mm} \noindent\textbf{Overview.} Given an input point cloud $P_0 \in \mathbb{R}^{N \times 3}$ (and color information $C_0 \in \mathbb{R}^{N \times 3}$ if available), we first apply Farthest Point Sampling (FPS), similar to previous works \cite{sparseconv,pointbert}, to obtain a set of $S$ center points, denoted as $P \in \mathbb{R}^{S \times 3}$. Next, kNN is applied to form a local patch with $k$ points around each center. These point patches (along with points' color) are then processed by a mini-PointNet \cite{pointnet} to obtain point tokens $E \in \mathbb{R}^{S \times C}$, where $C$ is the dimension of the token embedding space. An encoder composed of \( L \) DuoMamba blocks is employed to propagate information across local patches and capture global features. Finally, the encoder outputs are passed through a linear layer followed by average pooling to produce a single vector \( z^P \in \mathbb{R}^C \), which can be used for cross-modal contrastive learning as described in Section \ref{sec:occtip}. Figure \ref{fig:fig3_proposed_model} illustrates our network. 

\vspace{1mm} \noindent\textbf{DuoMamba Block.} At the core of the proposed architecture is the two-stream DuoMamba block, which leverages Mamba’s linear complexity \cite{mamba} for improved efficiency over the Transformer's quadratic self-attention \cite{pointbert,pointmae,point_trans_v1,pointtransformer}. We introduce two key adaptations to Mamba, originally designed for structured sequence data, to efficiently process point clouds.
\textbf{First}, we use Hilbert and Trans-Hilbert \cite{hilbert_curve} space-filling curves to transform 3D point clouds into 1D sequences. Unlike audio or text, point clouds are essentially sets of unordered 3D coordinates, making them challenging to process with order-aware models like Mamba. By sorting point tokens along Hilbert curves, adjacent patches in the sequence correspond to nearby regions in 3D space, facilitating local information propagation compared to random ordering. Additionally, using two Hilbert variants allows us to capture more diverse spatial relationships, enriching local point interactions \cite{point_trans_v3}. 
\textbf{Second}, we replace the causal 1D convolution used in previous Mamba-based models \cite{vmamba,mamba,pointmamba,mamba3d} with the standard convolution. In tasks like audio and language modeling where data follows a natural order, restricting tokens to attend only to preceding ones can be beneficial \cite{causal_1d_wavenet}. By contrast, for spatial data like point clouds, allowing patches to aggregate information bidirectionally along scanning curves enables them to consider neighbors in every direction, providing a more comprehensive spatial context.

Figure \ref{fig:fig3_proposed_model} illustrates our DuoMamba block, which consists of two parallel streams that extract point features using two S6 modules \cite{mamba}. In each branch, point patches are ordered along the Hilbert or Trans-Hilbert curve, then local relationships are propagated with a 1D convolution. S6 further facilitates information flow between tokens and models long-range dependencies. Finally, two sequences are reordered and combined to produce output. Specifically, the $l$-th block transforms the output $Z_{l-1}^{\text{out}}$ from the previous module as follows:

\vspace{-3mm}
{\footnotesize
\begin{equation}
\begin{aligned}
    & Z_l^{\text{in}} = \mathrm{LayerNorm} \left( Z_{l-1}^{\text{out}} \right), 
    & &Z_l = \mathrm{SiLU} \left( \mathrm{Linear} \left( Z_l^{\text{in}} \right) \right),\\
    & H'_l = \mathrm{HSort} \left( \mathrm{Linear} \left( Z_l^{\text{in}} \right) \right),     
    & &H''_l = \mathrm{SiLU} \left( \mathrm{Conv1D} \left( H'_l \right) \right),\\
    & T'_l = \mathrm{THSort} \left( \mathrm{Linear} \left( Z_l^{\text{in}} \right) \right),
    & &T''_l = \mathrm{SiLU} \left( \mathrm{Conv1D} \left( T'_l \right) \right),\\
    & H_l = \mathrm{Unsort} \left( \mathrm{S6} \left( H''_l \right) \right) \odot Z_l, 
    & &T_l = \mathrm{Unsort} \left( \mathrm{S6} \left( T''_l \right) \right) \odot Z_l,\\
    & Z_l^{\text{out}} = Z_{l-1}^{\text{out}} + \mathrm{Linear} \left( H_l + T_l \right),
\end{aligned}    
\end{equation}
}where $\mathrm{HSort}$ and $\mathrm{THSort}$ represent sorting operations based on Hilbert and Trans-Hilbert curves while $\mathrm{Unsort}$ is the operation that restores the original order.
\section{Experiments}
\label{sec:experiments}
\begin{table}[t!]
\centering
\small
    \setlength\aboverulesep{0pt}\setlength\belowrulesep{0pt}
    \setlength{\tabcolsep}{6.5pt}  
    \resizebox{0.48\textwidth}{!}{%
    \begin{tabular}{c|c|ccc|ccc}
    \toprule
    \multirow{2}{*}{Method} & \multirow{2}{*}{Encoder} & \multicolumn{3}{c|}{ModelNet40-P} & \multicolumn{3}{c}{ScanObjectNN} \\ \cline{3-8} 
              &          & Top 1 & Top 3 & Top 5 & Top 1 & Top 3 & Top 5 \\ 
    \midrule
    \midrule
    OpenShape \cite{openshape} & \multirow{5}{*}{SparseConv \cite{sparseconv}} &   42.1    &   61.6    &  69.4     &   52.7    &  72.7     &   83.6    \\ 
    TAMM \cite{tamm}  &   &   45.5    &   64.8    &   73.1    & 57.9 & 75.3 & 83.1 \\ 
    MixCon3D \cite{mixcon3d} &    &   -    &   -    &   -    &    54.4   &   73.9    &   83.3    \\ 
    MixCon3D$^\dagger$ \cite{mixcon3d} &   &   42.1    &  59.3     &   67.5    &   56.0   &   73.2    &   82.8    \\  
    \rowcolor{gray!20}
    OccTIP & \cellcolor{white} & \textbf{64.5} & \textbf{81.0} & \textbf{86.7} & \textbf{61.7} & \textbf{78.4} & \textbf{86.9}  \\ 
    \midrule
    OpenShape \cite{openshape} & \multirow{5}{*}{PointBERT \cite{pointbert}} &   46.3    &   64.2    &   71.9    &  51.3     &   69.4    &  78.4   \\  
    TAMM  \cite{tamm} &      &    45.6    &   66.2    &   74.7   & 54.8 & 74.5 & 83.3 \\     
    MixCon3D \cite{mixcon3d} &       &   -    &   -    &   -    &   52.6   &   69.9    &    78.7   \\
    MixCon3D$^\dagger$ \cite{mixcon3d} &    &   50.3    &   69.7    &   78.6    &   55.5   &   72.8    &    81.1   \\
    \rowcolor{gray!20}
    OccTIP & \cellcolor{white} & \textbf{67.7} & \textbf{82.7} & \textbf{87.3} & \textbf{60.6} & \textbf{78.2} & \textbf{86.0}  \\ 
    \midrule    
    OpenDlign \cite{opendlign} & ViT-H-14 \cite{vit} & - & - & - & 59.5 & 76.8 & 83.7 \\ 
    \rowcolor{gray!20}
    OccTIP & DuoMamba &   \textbf{67.7}    &   \textbf{82.9}    & \textbf{87.8} &  \textbf{63.5}    &  \textbf{81.3}     &   \textbf{89.2}    \\ 
    \bottomrule
    \end{tabular}
    }
\caption{Zero-shot classification accuracy on ModelNet-P and ScanObjectNN \cite{scanobjectnn}. ScanObjectNN results are from prior work, while ModelNet40-P results are obtained by running official pretrained models. $^\dagger$: As MixCon3D \cite{mixcon3d} weights are unavailable, we retrain 3D encoders using the authors' code, achieving higher accuracy than previously reported, which we use in all comparisons.}
\label{tab:zeroshot}    
\vspace{-5mm}
\end{table}

\begin{table}[h]
\centering
\small 
    \setlength\aboverulesep{0pt}\setlength\belowrulesep{0pt}
    \setlength{\tabcolsep}{6.5pt}  
    \resizebox{0.48\textwidth}{!}{%
    \begin{tabular}{c|c|ccccc}
    \toprule
    \multirow{2}{*}{Method} & \multirow{2}{*}{Encoder} & \multicolumn{5}{c}{ScanObjectNN} \\ \cline{3-7} 
              & & 1-shot & 2-shot & 4-shot & 8-shot & 16-shot \\ 
    \midrule
    \midrule
    OpenShape$^*$ \cite{openshape} & \multirow{4}{*}{SparseConv \cite{sparseconv}} & 41.7 & 49.1 & 58.1 & 63.4 & 70.0 \\ 
    MixCon3D$^\dagger$ \cite{mixcon3d} & & 42.4 & 51.1 & 59.9 & 65.4 & 71.5 \\
    TAMM$^*$ \cite{tamm} & & 43.6 & 52.8 & 59.5 & 70.9 & 73.4 \\ 
    \rowcolor{gray!20}
    OccTIP & \cellcolor{white} & \textbf{48.5} & \textbf{58.2} & \textbf{67.3} & \textbf{72.3} & \textbf{76.7} \\  
    \midrule
    OpenShape$^*$ \cite{openshape} & \multirow{4}{*}{PointBERT \cite{pointbert}} & 38.4 & 50.9 & 61.3 & 67.7 & 71.5 \\  
    MixCon3D$^\dagger$ \cite{mixcon3d} & & 39.0 & 53.9 & 61.2 & 68.0 & 71.5 \\    
    TAMM$^*$ \cite{tamm} & & 46.3 & 56.7 & 66.6 & {73.4} & \textbf{77.6} \\ 
    \rowcolor{gray!20}
    OccTIP & \cellcolor{white} & \textbf{50.8} & \textbf{60.8} & \textbf{68.6} & \textbf{73.7} & 76.8 \\ 
    \midrule
    \rowcolor{gray!20}
    OccTIP & DuoMamba & \textbf{52.5} & \textbf{63.0} & \textbf{70.2} & \textbf{76.0} & \textbf{79.4} \\      
    \bottomrule
    \end{tabular}
    }
\caption{Few-shot linear probing accuracy on ScanObjectNN \cite{scanobjectnn}. (*: results obtained using released pretrained weights, $^\dagger$: results reproduced using the authors' public code.)} %
\label{tab:fewshot}  
\vspace{-5mm}
\end{table}

We generate text-image-point cloud pretraining triplets using 52,417 3D models from the ShapeNetCore \cite{shapenet} dataset following the procedure in Section \ref{sec:occtip}. For evaluation, we create 12 partial point clouds for each ModelNet40 \cite{modelnet40} test object, resulting in ModelNet40-P with 29,610 occluded point clouds of 40 classes. We also use three other real-scanned benchmarks: ScanObjectNN \cite{scanobjectnn}, ScanNetV2 \cite{scannetv2}, and SUN RGB-D \cite{sunrgbd}. For fair comparisons, we mainly evaluate our work against methods that are also pretrained on ShapeNetCore \cite{shapenet} objects, including OpenShape \cite{openshape}, MixCon3D \cite{mixcon3d}, and TAMM \cite{tamm}. 

\vspace{1mm} \noindent\textbf{Evaluation Tasks.} We conduct extensive experiments in four recognition tasks with varying difficulty levels (zero-shot classification, few-shot linear probing, real-world instance recognition, and zero-shot object detection) to demonstrate the superiority of our pretraining framework OccTIP and the proposed architecture DuoMamba. The details of each experiment will be described in the following subsections.

\vspace{1mm}
\noindent\textbf{Implementation and Training Details.} We implement our method in PyTorch \cite{pytorch} and conduct all experiments on a single NVIDIA RTX 4090 GPU. We sample 2,048 points per point cloud as input and train the 3D encoders for 200 epochs using AdamW \cite{adamw} with a 10-epoch warmup, which takes around 1.5 days. Following prior works \cite{openshape,mixcon3d,tamm}, we use OpenCLIP ViT-bigG-14 \cite{openclip} as pretrained image-text encoders. Further details are in the supplementary material.

\subsection{Zero-Shot Classification}
\begin{table*}[t]
\centering
\small
    \setlength\aboverulesep{0pt}\setlength\belowrulesep{0pt}
    \setlength{\tabcolsep}{6.5pt}  
    \resizebox{\textwidth}{!}{%
    \begin{tabular}{l|c|ccccccccccccccccc}
    \toprule
    Method & Avg. & Cab & Bed & Chair & Sofa & Tabl & Door & Wind & Bksf & Pic & Cntr & Desk & Curt & Fridg & ShwrCurt & Toil & Sink & Bath \\
    \midrule
    \midrule
    PointCLIP \cite{pointclip} w/ TP. & 26.1 & 0.0 & 55.7 & 72.8 & 5.0 & 5.1 & 1.7 & 0.0 & 77.2 & 0.0 & 0.0 & 51.7 & 0.0 & 0.0 & 40.3 & 85.3 & 49.2 & 0.0 \\
    CLIP2Point \cite{clip2point} w/ TP.  & 35.2 & 11.8 & 0.0 & 45.1 & 27.6 & 10.5 & 61.5 & 2.6 & 71.9 & 0.3 & 33.6 & 29.9 & 4.7 & 11.5 & 92.4 & 86.1 & 34.0 & 72.2 \\
    CLIP$^2$ \cite{clip2} & 38.5 & 67.2 & 32.6 & 69.3 & 42.3 & 18.3 & 19.1 & 4.0 & 62.6 & 1.4 & 12.7 & 52.8 & 40.1 & 9.1 & 41.0 & 71.0 & 45.5 & 59.7 \\
    \midrule
    OpenShape$^*$ \cite{openshape} (SparseConv) & 39.9 & 0.0 & 59.3 & 76.8 & 61.9 & 42.3 & 57.0 & 14.2 & 71.4 & 31.1 & 0.0 & 67.7 & 20.9 & 0.0 & 0.0 & 89.7 & 43.9 & 41.9 \\ 
    MixCon3D$^\dagger$ \cite{mixcon3d} (SparseConv) & 39.9 & 0.0 & 69.1 & 69.6 & 67.0 & 43.7 & 51.0 & 5.7 & 75.3 & 53.2 & 1.9 & 59.8 & 4.5 & 0.0 & 10.7 & 93.1 & 12.2 & 61.3 \\ 
    TAMM$^*$ \cite{tamm} (SparseConv) & 43.7 & 0.5 & 67.9 & 72.5 & 72.2 & 52.9 & 51.8 & 20.2 & 77.9 & 50.0 & 25.0 & 61.4 & 7.5 & 0.0 & 0.0 & 87.9 & 36.7 & 58.1 \\
    \rowcolor{gray!20}
    OccTIP (SparseConv) & \textbf{45.3} & 1.1 & 71.6 & 80.7 & 87.6 & 45.7 & 52.3 & 5.0 & 70.1 & 56.3 & 3.9 & 64.6 & 4.5 & 1.8 & 0.0 & 96.6 & 51.0 & 77.4 \\ 
    \midrule
    OpenShape$^*$ \cite{openshape} (PointBERT) & 40.5 & 0.5 & 60.5 & 70.5 & 67.0 & 41.7 & 50.1 & 9.2 & 72.7 & 44.1 & 3.9 & 75.6 & 7.5 & 0.0 & 0.0 & 72.4 & 54.1 & 58.1 \\    
    MixCon3D$^\dagger$ \cite{mixcon3d} (PointBERT) & 40.3 & 0.8 & 60.5 & 73.2 & 74.2 & 56.9 & 65.5 & 2.5 & 64.9 & 61.7 & 1.9 & 62.2 & 3.0 & 0.0 & 10.7 & 67.2 & 8.2 & 71.0 \\ 
    TAMM$^*$ \cite{tamm} (PointBERT) & 38.6 & 1.9 & 56.8 & 71.1 & 66.0 & 46.9 & 65.7 & 17.7 & 67.5 & 23.4 & 7.7 & 75.6 & 0.0 & 0.0 & 0.0 & 72.4 & 44.9 & 38.7 \\   \rowcolor{gray!20}
    OccTIP (PointBERT) & \textbf{47.8} & 11.6 & 80.3 & 73.0 & 83.5 & 54.9 & 56.8 & 16.7 & 61.0 & 72.5 & 1.9 & 53.5 & 11.9 & 21.1 & 7.1 & 91.4 & 41.8 & 71.2 \\ 
    \midrule
    \rowcolor{gray!20}
    OccTIP (DuoMamba) & \textbf{49.0} & 4.6 & 79.0 & 77.1 & 87.6 & 54.6 & 52.3 & 10.3 & 79.2 & 61.3 & 3.9 & 53.5 & 31.3 & 16.8 & 0.0 & 94.8 & 56.1 & 71.0 \\
    \bottomrule
    \end{tabular}%
    }
\caption{Zero-shot classification accuracy on the real-world ScanNetV2 \cite{scannetv2} instances. (*: results obtained using released pretrained weights, $^\dagger$: results reproduced using the authors' public code.)}
\label{tab:zeroshot_scannet}    
\vspace{-5mm}
\end{table*}
A pretrained network can perform zero-shot classification without fine-tuning by comparing its 3D shape representations to text embeddings of candidate categories. To assess the quality of the learned latent space, we conduct zero-shot classification experiments on ModelNet40-P and ScanObjectNN \cite{scanobjectnn} (OBJ\_ONLY version). ScanObjectNN \cite{scanobjectnn} contains 2,890 real-scanned point clouds in 15 classes, providing a more realistic benchmark than our synthetic ModelNet40-P. 
As summarized in Table \ref{tab:zeroshot}, our method significantly outperforms previous approaches. For SparseConv \cite{sparseconv} and PointBERT \cite{pointbert}, our framework improves their performance by 19.0\% and 17.4\% on ModelNet40-P compared to the best existing results. On ScanObjectNN \cite{scanobjectnn}, OccTIP raises accuracy by 3.8\% and 5.1\%, reaching 61.7\% and 60.6\%, both surpassing the current state-of-the-art OpenDlign \cite{opendlign}. These results highlight our framework's effectiveness in bridging the training-testing domain gap for improved real-world recognition. Furthermore, when combining OccTIP with DuoMamba, accuracy increases by an additional 1.8\%, establishing a new state-of-the-art of 63.5\% on ScanObjectNN \cite{scanobjectnn} and demonstrating DuoMamba's learning prowess in cross-modal representation learning.
\subsection{Few-Shot Linear Probing}
\label{main_exp_fewshot}
To further evaluate the learned embedding space, we conduct few-shot linear probing on ScanObjectNN \cite{scanobjectnn}. Following OpenShape \cite{openshape}, we use a pretrained model to extract features for all test samples and train a linear classifier using only a limited number of labeled instances per class. We report classification accuracy across a range of few-shot settings, specifically with 1, 2, 4, 8, and 16 labeled samples per category. As shown in Table \ref{tab:fewshot}, when trained with OccTIP, PointBERT \cite{pointbert} and SparseConv \cite{sparseconv} consistently achieve better results than existing approaches in nearly all few-shot settings. DuoMamba further enhances performance, attaining the highest accuracy under all configurations.  
This showcases the proposed network's strong learning capacity and highlights our framework's effectiveness in facilitating transferable feature learning, underscoring its applicability in label-scarce scenarios.
\subsection{Real-World Instance Recognition}
\label{subsec:ins_cls}
Following prior work \cite{clip2, tamm}, we test the pretrained models' capability to understand complex objects with the real-world instance recognition task. In this setting, the models have to classify object instances from a scene in a zero-shot manner. Using the same setting as CLIP$^2$ \cite{clip2}, we report results on the popular scene-level ScanNetV2 \cite{scannetv2} dataset. We extract object instances using ground-truth instance masks and classify them with the pretrained models. Table \ref{tab:zeroshot_scannet} summarizes the per-class accuracy and overall class average.

Our method significantly outperforms approaches pretrained on 1.6M real-world text-image-point cloud triplets, including PointCLIP w/TP \cite{pointclip}, CLIP2Point w/TP \cite{clip2point}, and CLIP$^2$ \cite{clip2}. Compared to other ShapeNetCore-based pretraining methods, OccTIP consistently boosts PointBERT \cite{pointbert} and SparseConv \cite{sparseconv} accuracy by 7.3\% and 1.6\%, respectively. When combined with DuoMamba, the class-average accuracy rises by an additional 1.2\%, reaching 49.0\%. These results once again underscore our model’s strong learning capacity and highlight the effectiveness of our pretraining framework for robust feature extraction in real-world 3D shape understanding.
\begin{table}[!h]
\centering
\small
    \setlength\aboverulesep{0pt}\setlength\belowrulesep{0pt}
    \setlength{\tabcolsep}{6.5pt}  
    \resizebox{0.4\textwidth}{!}{%
    \begin{tabular}{c|c|cc} 
    \toprule
    & Method & ScanNetV2 & SUN RGB-D \\ 
    \midrule
    \midrule
    \multirow{6}{*}{mAP$_{25}$} & PointCLIP \cite{clip2point} & 6.0 & - \\ 
    & PointCLIP V2 \cite{pointclipv2} & 19.0 & - \\
    & OpenShape$^*$ \cite{openshape} & 20.4 & 18.6 \\ 
    & MixCon3D$^\dagger$ \cite{mixcon3d} & 24.1 & 18.7 \\ 
    & TAMM$^*$ \cite{tamm} & 23.1 & 18.9 \\
    \rowcolor{gray!20}
    \cellcolor{white} & OccTIP & \textbf{28.9} & \textbf{24.4} \\ 
    \midrule
    \multirow{6}{*}{mAP$_{50}$} & PointCLIP \cite{pointclip} & 4.8 & - \\ 
    & PointCLIP V2 \cite{pointclipv2} & 11.5 & -  \\ 
    & OpenShape$^*$ \cite{openshape} & 16.1 & 9.8 \\ 
    & MixCon3D$^\dagger$ \cite{mixcon3d} & 19.1 & 9.6 \\ 
    & TAMM$^*$ \cite{tamm} & 18.1 & 10.0 \\
    \rowcolor{gray!20}
    \cellcolor{white} & OccTIP & \textbf{22.7} & \textbf{13.0}  \\
    \bottomrule
    \end{tabular}
    }
\caption{Zero-shot 3D object detection results on ScanNetV2 \cite{scannetv2} and SUN RGB-D \cite{sunrgbd}. For complete results, please refer to our supplementary materials. (*: results obtained using released pretrained weights, $^\dagger$: results reproduced using the authors' code.)
}
\label{tab:obj_det}    
\vspace{-3mm}
\end{table}

\subsection{Zero-Shot 3D Object Detection}
\label{subsec:obj_det}
{
\begin{table*}[t]
\centering
\small
\setlength\aboverulesep{0pt}\setlength\belowrulesep{0pt}
\setlength{\tabcolsep}{6.5pt}  
\begin{subtable}[b]{0.31\textwidth}
\centering
\resizebox{\textwidth}{!}{%
    \begin{tabular}{c|ccc|cc} 
    Setting & Hilbert & Trans-Hilbert & Conv1D & ScanObjectNN \\ 
    \midrule
    \midrule
    \texttt{(i)} & - & - & - &  60.6 \\ 
    \midrule
    \texttt{(ii)} & \cmark & - & \cmark &  62.2 \\ 
    \texttt{(iii)} & - & \cmark & \cmark & 61.7 \\ 
    \midrule
    \texttt{(iv)} & \cmark & \cmark & - & 63.1 \\
    \rowcolor{gray!20}
    \texttt{(v)} & \cmark & \cmark & \cmark & \textbf{63.5} \\         
    \end{tabular}
    }
\caption{Contribution of each component in our two-stream DuoMamba block.}
\label{tab:ablation_component}
\end{subtable}
\hfill
\begin{subtable}[b]{0.26\textwidth}
\centering
\resizebox{\textwidth}{!}{%
    \begin{tabular}{l|cc} 
    Point order & ScanObjectNN \\ 
    \midrule
    \midrule
    FPS order & 61.8 \\
    Z-order and Trans-Z-order & 62.7 \\  
    Hilbert and Z-order & 62.4 \\
    \rowcolor{gray!20}
    Hilbert and Trans-Hilbert & \textbf{63.5} \\ 
    \end{tabular}
    }
\caption{Effect of different sorting strategies on DuoMamba's performance.}
\label{tab:ablation_curves}
\end{subtable}
\hfill
\begin{subtable}[b]{0.39\textwidth}
\centering
\resizebox{\textwidth}{!}{%
\begin{tabular}{l|cc|c} 
    Model & Param. (M) $\downarrow$  & FLOPs (G) $\downarrow$ & ScanObjectNN $\uparrow$ \\ 
    \midrule
    \midrule
    Mamba3D \cite{mamba3d} & 29.9 & 6.8 & 60.7  \\ 
    PointMamba \cite{pointmamba} & 21.4 & 10.3 & 62.6 \\ 
    \rowcolor{gray!20}
    DuoMamba & 29.2 & 7.1 & \textbf{63.5} \\ 
    \end{tabular}
    }
\caption{Mamba-based encoders comparison. Our DuoMamba architecture achieves a better computation-performance trade-off than previous methods.}
\label{tab:mamba_comparison}
\end{subtable}
\vspace{-2mm}
\caption{Ablation studies to validate the design of our proposed network and comparisons with existing Mamba-based point cloud models.}
\label{tab:ablation}
\vspace{-3mm}
\end{table*}
}

To showcase how our pretrained model can be combined with existing methods to tackle more challenging tasks, we conduct zero-shot 3D object detection experiments on ScanNetV2 \cite{scannetv2} and SUN RGB-D \cite{sunrgbd}. Following the setup in PointCLIP V2 \cite{pointclipv2}, we leverage 3DETR-m \cite{3detr} detector to predict 3D bounding boxes, which enables the extraction of points corresponding to each object instance. Our pretrained 3D network is then applied to classify these object point clouds in a zero-shot manner. Based on 3DETR-m’s localization and our classifier’s semantic predictions, we calculate the mean Average Precision (mAP) at IoU thresholds of 0.25 and 0.5 across 18 object categories in ScanNetV2 and 10 most frequent classes in SUN RGB-D.

As shown in Table \ref{tab:obj_det}, our method achieves mAP$_{25}$ and mAP$_{50}$ scores of 28.9\% and 22.7\% on ScanNetV2, marking significant improvements of 9.9\% and 11.2\% over the depth-based PointCLIP V2 \cite{pointclipv2}. Compared to other point-based methods, we outperform the second-best approach MixCon3D \cite{mixcon3d} by 4.8\% and 3.6\% on mAP$_{25}$ and mAP$_{50}$, respectively. A similar trend is observed on the SUN RGB-D benchmark, where our approach achieves the highest mAP$_{25}$ and mAP$_{50}$ scores of 24.4\% and 13.0\%.
The results again confirm the superiority of our method in learning robust features for recognizing noisy 3D objects in complex scenes, highlighting its strong potential for general 3D open-world learning.

\subsection{Visualization of the Embedding Space}
We further compare the latent spaces of our model with those of existing works. Specifically, we use the pretrained encoders to extract features of ScanObjectNN \cite{scanobjectnn} test instances and employ t-SNE \cite{tsne} for dimensionality reduction. As shown in Figure \ref{fig:tsne}, our method exhibits superior separation and clustering of object classes compared to previous approaches. Note that the pretrained model did not encounter any of these samples during training, yet it successfully captures the characteristics of each category and minimizes overlap between them. This separation indicates that our method’s feature representations are more robust, leading to better real-world zero-shot performance as demonstrated in previous experiments. 
\subsection{Ablation Study}
\label{subsec:ablation}
We conduct ablation studies and report the zero-shot classification accuracy on ScanObjectNN \cite{scanobjectnn} to validate the design of DuoMamba.
We also compare with two existing Mamba-based models to demonstrate the advantages of our proposed architecture.

\vspace{1mm} 
\noindent\textbf{Component Contribution.} We analyze the impact of different components in our DuoMamba block, with results summarized in Table \ref{tab:ablation_component}. In the baseline setting \texttt{(i)}, applying the original Mamba \cite{mamba} to FPS-based ordered sequences yields the lowest accuracy of 60.6\%. Replacing causal 1D convolutions with the standard ones and using either Hilbert \texttt{(ii)} or Trans-Hilbert \texttt{(iii)} ordering consistently improves the performance, with higher accuracy from Hilbert order. Combining both curves with standard 1D convolution as in DuoMamba \texttt{(v)} leads to the best accuracy of 63.5\%. Without the standard 1D convolutions as in \texttt{(iv)}, accuracy drops 0.4\% to 63.1\%. These findings emphasize the importance of integrating geometric structures from both Hilbert curves with standard 1D convolutions for optimal information propagation.

\vspace{1mm} 
\noindent\textbf{The Effect of Scanning Routines.} We further explore the impact of scanning patterns for serializing point clouds and report the results in Table \ref{tab:ablation_curves}. We compare the performance of the default FPS order with three combinations of the widely used Z-order and Hilbert curves \cite{hilbert_curve}. Our results show that combining two variants of Z-order outperforms FPS, and using two Hilbert curves achieves the highest accuracy. This improvement is attributed to the fact that space-filling curves better preserve spatial relationships between point patches, enhancing information flow among nearby tokens in the sequence. Moreover, the superior locality-preserving properties of Hilbert curves over Z-order \cite{point_trans_v3} contribute to a performance boost when used for processing point cloud sequences, as implemented in our DuoMamba block.

\vspace{1mm} 
\noindent\textbf{Mamba-Based Encoder Comparison.} 
To justify the significance of our new architecture, we compare it with two existing Mamba-based models: Mamba3D \cite{mamba3d} and PointMamba \cite{pointmamba}. Table \ref{tab:mamba_comparison} shows that DuoMamba surpasses both models on the ScanObjectNN \cite{scanobjectnn} benchmark, outperforming Mamba3D \cite{mamba3d} by a significant margin of 2.8\% in accuracy. Although DuoMamba has more parameters than PointMamba \cite{pointmamba}, it achieves better performance while also maintaining a lower FLOPs count\footnote{PointMamba's FLOPs is computed when using PyTorch's standard implementation for causal conv1D.}. Overall, the proposed architecture demonstrates a better computation-performance balance than both existing networks.
\vspace{-3mm}
\section{Conclusion}
\label{sec:conclusion}
In this paper, we propose an occlusion-aware multi-modal pretraining framework for open-world 3D shape recognition. Our method uses synthetic 3D models to generate partial point clouds for pretraining, effectively reducing the training-testing domain gap and enhancing real-world recognition performance. Moreover, we introduce a Mamba-based architecture for point cloud processing, offering better performance with lower FLOPs and latency than Transformer-based networks. We hope our paper paves the way for future research on more realistic pretraining and computationally efficient models.

\vspace{1mm}
\noindent\textbf{Limitations.} Due to resource constraints, we have not been able to leverage Objaverse \cite{objaverse} - the largest dataset with nearly 800K 3D objects - for pretraining, which we believe could further enrich the learned latent space and enhance recognition performance. 

\paragraph{Acknowledgment.} The authors sincerely thank Mr. Trong-Thang Pham for his valuable feedback during the preparation of this paper. This research was supported by the Australian Research Council (ARC) under discovery grant project \#240101926. Professor Ajmal Mian is the recipient of an ARC  Future Fellowship Award (project \#FT210100268) funded by the Australian Government. Mr. Khanh Nguyen is supported by the tuition fee scholarship from the University of Western Australia and a stipend from the ARC project \#FT210100268.

{
    \small
    \bibliographystyle{ieeenat_fullname}
    \bibliography{main}
}
\newpage 
\clearpage
\setcounter{figure}{4}
\setcounter{table}{5}
\setcounter{section}{7}

\maketitlesupplementary
\section{Additional Discussion on Existing Works}
\paragraph{Discussion on Occlusion Methods.} \citet{modelnetc} simplified occlusion by treating it as a form of corruption, referred to as ``Drop Local,'' where k-NN clusters are randomly removed from point clouds. They then proposed an architecture and an augmentation strategy (based on deforming and mixing objects) to address \textit{general} corruptions rather than focusing on occlusion. \citet{mvtn} introduced a viewpoint prediction module as a component for multi-view 3D recognition (which rely on 3D-to-2D projection). By predicting `good' views to render images from point clouds, indirectly, the recognition model becomes more robust to occlusion (empirically simulated by randomly cropping the object point clouds along canonical directions). In contrast, our OccTIP method more realistically simulates self-occlusion through the rendering process and integrates single-view point clouds during pretraining, improving occlusion robustness for \textit{any} point cloud encoders.

\vspace{-5mm}
\paragraph{Comparison with VisionMamba (Vim).} While Vim \cite{vim} also has a two-stream design, it has two key limitations: (1) reliance on one-directional neighborhood aggregation (CausalConv1D) and (2) only able to utilize a \textit{single} neighborhood structure due to its simple forward and backward scanning strategy. In contrast, DuoMamba uses Conv1D for bidirectional local aggregation and can flexibly process two diverse orderings (e.g., Hilbert, Trans-Hilbert) simultaneously within a single block to fully exploit 3D geometry of the point clouds. These technical enhancements lead to improved performance as shown in Table \ref{tab:rebuttal_vim_duomamba}.
\begin{table}[h!]
\centering
\setlength\aboverulesep{0pt}\setlength\belowrulesep{0pt}
    \setlength{\tabcolsep}{6.5pt}  
    \resizebox{0.4\textwidth}{!}{
    \begin{tabular}{c|cc|c}
        \toprule
        Dataset & Vim \cite{vim} & Vim \cite{vim} + Hilbert & \cellcolor{gray!20} DuoMamba  \\
        \midrule
        ModelNet40-P &  65.3 & 63.8 & \cellcolor{gray!20}\textbf{67.7} \\ 
        ScanObjectNN &  61.1 & {62.7} & \cellcolor{gray!20}\textbf{63.5} \\
        \bottomrule
    \end{tabular}
    }
\caption{Zero-shot accuracy of Vim and DuoMamba.}
\label{tab:rebuttal_vim_duomamba}
\end{table}

\vspace{-5mm}
\section{Implementation Details}
\paragraph{Triplet Generations.} We render RBG images with a resolution of $512 \times 512$ and a transparent background. Similar to OpenShape \cite{openshape}, descriptions for each object come from three sources: (1) raw texts from the dataset's metadata, (2) captions generated by BLIP \cite{blip} and Azure Cognitive Services, (3) retrieved captions from visually similar images in the LAION-5B \cite{laion_5b} dataset. The first source of captions (created from metadata) includes three texts: (a) object name, (b) object category, and (c) concatenation of the subcategory name. 

\paragraph{Training Details.} During pretraining, we use a batch size of 32 and randomly replace point colors with a constant value of 0.4 with a probability of 0.5. During testing, we assign the same constant value to point clouds that do not have color information, such as those in the ScanObjectNN \cite{scanobjectnn} dataset. For more efficient training, we precompute and cache text and image features from CLIP \cite{2dclip} and directly use them as inputs to the text and image projection heads. Since there is significant fluctuation when training with partial point clouds, we follow \cite{mixcon3d} to employ Exponential Moving Average (EMA) \cite{ema} with a decay factor of 0.9995 to stabilize the training process. We use a cosine learning rate scheduler with a base learning rate of $7\mathrm{e}{-4}$.

\begin{table*}[h]
\centering
    \setlength\aboverulesep{0pt}\setlength\belowrulesep{0pt}
    \setlength{\tabcolsep}{6.5pt}  
    \resizebox{\textwidth}{!}{%
    \begin{tabular}{c|c|c|cccccccccccccccccc} 
    \toprule
    & Method & Mean & Cab & Bed & Chair & Sofa & Tabl & Door & Wind 
    & Bksf & Pic & Cntr & Desk & Curt & Fridg & ShwrCurt & Toil & Sink & Bath & Bin \\ 
    \midrule
    \midrule
    \multirow{4}{*}{AP$_{25}$} & PointCLIP \cite{clip2point} & 6.00 & 3.99 & 4.82 & 45.16 & 4.82 & 7.36 & 4.62 & 2.19 & - & - & 1.02 & 4.00 & - & - & - & - & 13.40 & 6.46 & - \\ 
    & PointCLIP V2 \cite{pointclipv2} & 18.97 & \textbf{19.32} & 20.98 & 61.89 & 15.55 & 23.78 & 13.22 & \textbf{17.42} & - & - & \textbf{12.43} & 21.43 & - & - & - & - & 14.54 & 16.77 & - \\
    & OpenShape$^*$ \cite{openshape}  & 20.40 & 9.63 & 38.62 & 73.05 & 57.28 & 37.00 & 29.52 & 5.74 & 23.94 & 2.07 & \underline{3.37} & 16.25 & 1.25 & 4.45 & 0.84 & 9.00 & 22.76 & 16.21 & 16.23 \\ 
    & MixCon3D$^\dagger$ \cite{mixcon3d}  & \underline{24.11} & 11.55 & \underline{43.21} & \underline{79.33} & \underline{63.97} & \textbf{42.91} & 29.94 & 4.85 & \underline{25.26} & \textbf{3.98} & 1.49 & \underline{25.58} & 2.00 & \underline{4.95} & 0.81 & 13.23 & 20.58 & 38.03 & \underline{22.25} \\ 
    & TAMM$^*$ \cite{tamm}  & 23.07 & 10.03 & 32.68 & 75.16 & 55.73 & 36.72 & \underline{32.44} & 5.26 & 24.82 & 2.52 & 2.04 & 22.53 & \underline{2.11} & 3.26 & \underline{1.23} & \underline{17.83} & \underline{23.87} & \textbf{46.50} & 20.48 \\ 
    \rowcolor{gray!20}    
    & OccTIP  & \textbf{28.92} & \underline{12.85} & \textbf{56.43} & \textbf{80.41} & \textbf{68.78} & \underline{40.11} & \textbf{37.68} & \underline{7.09} & \textbf{30.51} & \underline{3.21} & {2.46} & \textbf{31.55} & \textbf{5.18} & \textbf{8.54} & \textbf{2.14} & \textbf{29.89} & \textbf{35.64} & \underline{41.93} & \textbf{26.24} \\ 
    \midrule
    \multirow{4}{*}{AP$_{50}$} & PointCLIP \cite{pointclip} & 4.76 & 1.67 & 4.33 & 39.53 & 3.65 & 5.97 & 2.61 & 0.52 & - & - & 0.42 & 2.45 & - & - & - & - & 5.27 & 1.31 & - \\ 
    & PointCLIP V2 \cite{pointclipv2} & 11.53 & \textbf{10.43} & 13.54 & 41.23 & 6.60 & 15.21 & 6.23 & \textbf{11.35} & - & - & \textbf{6.23} & 10.84 & - & - & - & - & \underline{11.43} & 10.14 & - \\ 
    & OpenShape$^*$ \cite{openshape}  & 16.12 & 3.78 & 36.99 & 62.48 & 49.48 & 33.05 & 17.40 & 2.12 & 21.97 & 0.61 & \underline{1.34} & 11.97 & 0.45 & 4.18 & 0.59 & 8.38 & 10.68 & 16.16 & 8.55 \\ 
    & MixCon3D$^\dagger$ \cite{mixcon3d}  & \underline{19.09} & 3.61 & \underline{41.90} & \underline{67.67} & \underline{51.13} & \textbf{38.22} & 17.34 & 1.56 & \underline{23.44} & \textbf{1.56} & 0.36 & \underline{18.63} & 0.59 & \underline{4.71} & 0.43 & 12.07 & 9.18 & 37.69 & \underline{13.51} \\ 
    & TAMM$^*$ \cite{tamm}  & 18.11 & 3.10 & 31.64 & 64.35 & 42.51 & 30.82 & \underline{20.55} & 2.11 & 21.26 & 0.85 & 0.50 & 17.71 & \underline{0.80} & 3.09 & \underline{0.81} & \underline{17.00} & 10.44 & \textbf{46.27} & 12.26 \\
    \rowcolor{gray!20}    
    & OccTIP  & \textbf{22.73} & \underline{5.44} & \textbf{54.77} & \textbf{68.91} & \textbf{55.53} & \underline{34.55} & \textbf{22.55} & \underline{2.92} & \textbf{25.71} & \underline{0.98} & 0.84 & \textbf{22.91} & \textbf{2.34} & \textbf{8.36} & \textbf{1.31} & \textbf{27.27} & \textbf{16.86} & \underline{41.65} & \textbf{16.27} \\   
    \bottomrule
    \end{tabular}
    }
\caption{Zero-shot 3D object detection results on ScanNetV2 \cite{scannetv2}. Our method OccTIP achieves the highest mAP and consistently has the highest or second-highest AP scores across most categories, showing the superiority of the proposed approach in complex real-world recognition. (*: results obtained using released pretrained weights, $^\dagger$: results reproduced using the authors' public code.)}
\label{tab:supp_obj_det_scannetv2}    
\end{table*}

\begin{table*}[t]
\centering
    \setlength\aboverulesep{0pt}\setlength\belowrulesep{0pt}
    \setlength{\tabcolsep}{6.5pt}  
    \resizebox{0.8\textwidth}{!}{%
    \begin{tabular}{c|c|c|cccccccccc} 
    \toprule
    & Method & Mean & Bed & Table & Sofa  & Chair & Toilet & Desk  & Dresser & Night Stand & Bookshelf & Bathtub \\
    \midrule
    \midrule
    \multirow{4}{*}{AP$_{25}$} & OpenShape$^*$ \cite{openshape} & 18.61 & \underline{33.09} & 24.18 & 28.96 & 45.51 & 10.42 & 13.58 & \underline{2.75} & \textbf{11.77} & 11.13 & 4.71 \\ 
    & MixCon3D$^\dagger$ \cite{mixcon3d} & 18.69 & 28.25 & 26.75 & \textbf{34.44} & \underline{47.77} & 6.05 & \underline{15.76} & 2.31 & \underline{11.56} & 6.91 & 7.14 \\ 
    & TAMM$^*$ \cite{tamm} & \underline{18.91} & 18.15 & \underline{27.78} & 27.67 & 47.00 & \textbf{21.41} & 14.54 & 2.43 & 10.81 & \underline{11.14} & \underline{8.20} \\
    \rowcolor{gray!20}
    & OccTIP  & \textbf{24.37} & \textbf{43.45} & \textbf{29.21} & \underline{34.22} & \textbf{51.19} & \underline{12.78}  & \textbf{18.16} & \textbf{3.76} & 11.14       & \textbf{13.96}     & \textbf{25.90}   \\ 
    \midrule
    \multirow{4}{*}{AP$_{50}$} & OpenShape$^*$ \cite{openshape} & 9.78 & \underline{23.71} & 9.01 & 20.85 & 24.37 & 7.74 & 3.02 & \underline{1.00} & \underline{5.47} & \underline{1.77} & 0.89 \\ 
    & MixCon3D$^\dagger$ \cite{mixcon3d} &  9.63 & 17.97 & 10.22 & \underline{24.53} & \underline{26.00} & 3.80 & \underline{3.38} & 0.51 & \textbf{6.30} & 1.73 & 1.86 \\ 
    & TAMM$^*$ \cite{tamm} & \underline{9.96} & 12.37 & \underline{11.01} & 20.36 & 25.41 & \textbf{17.96} & 3.22 & 0.81 & 4.87 & 1.71 & \underline{1.90} \\
    \rowcolor{gray!20}
    & OccTIP  & \textbf{13.01} & \textbf{32.67} & \textbf{11.21} & \textbf{25.46} & \textbf{28.04} & \underline{8.50}   & \textbf{4.33}  & \textbf{1.71}    & 5.11        & \textbf{1.92}      & \textbf{11.18}   \\
    \bottomrule
    \end{tabular}
    }
\caption{Zero-shot 3D object detection results on SUN RGB-D \cite{sunrgbd}. Our method OccTIP achieves the highest mAP and consistently has the highest or second-highest AP scores across most categories, showing the superiority of the proposed approach in complex real-world recognition. (*: results obtained using released pretrained weights, $^\dagger$: results reproduced using the authors' public code.)}
\label{tab:supp_obj_det_sunrgbd}    
\end{table*}

\section{Comparisons with Previous Works Pretrained on Larger Datasets}
We further compare our method (pretrained on 52K ShapeNetCore \cite{shapenet} objects) with previous works pretrained on a significantly larger ensemble of 880K 3D objects from four datasets: ShapeNetCore \cite{shapenet}, ABO \cite{abo}, 3D-FUTURE \cite{3dfuture}, and Objaverse \cite{objaverse}. We use the official results reported in previous papers and evaluate all approaches on the real-world ScanObjectNN \cite{scanobjectnn} dataset to assess their recognition performance in practical scenarios.

\begin{figure}[!h]
    \centering
    \includegraphics[width=0.8\linewidth]{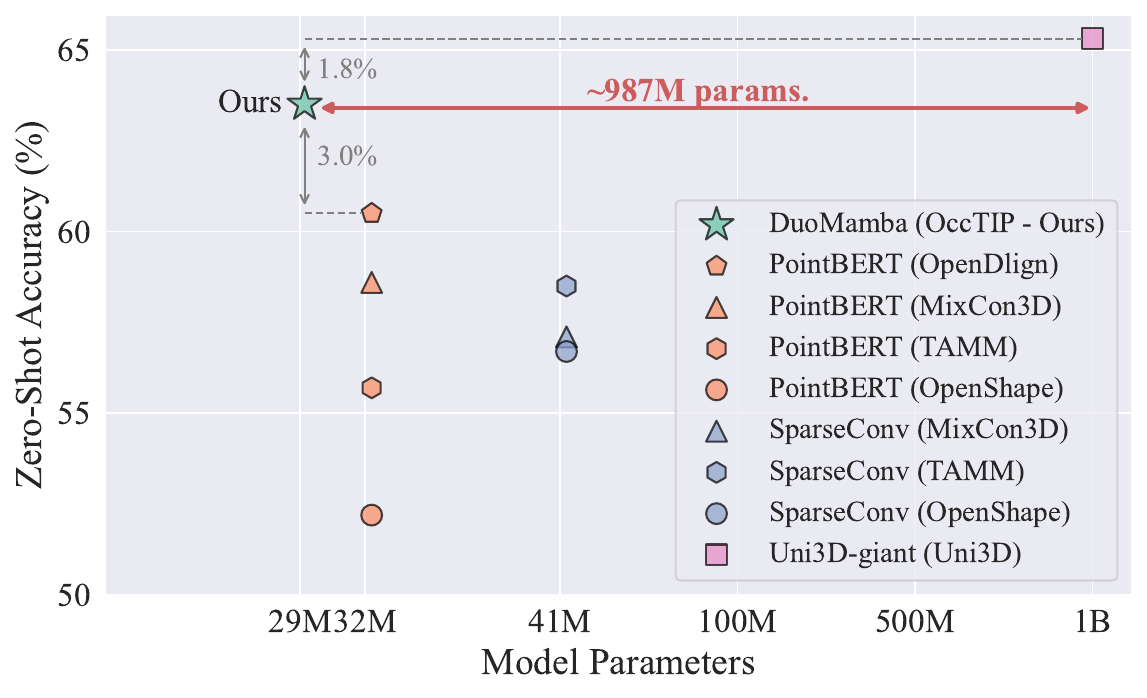}
    \vspace{-3mm}
    \caption{\textbf{Comparisons of model size and zero-shot accuracy on ScanObjectNN} \cite{scanobjectnn}.
    Our model is pretrained on 52K ShapeNetCore \cite{shapenet} objects, whereas all other approaches are pretrained on an ensemble of 880K objects from four datasets: Objaverse \cite{objaverse}, ABO \cite{abo}, 3D-FUTURE \cite{3dfuture}, and ShapeNetCore \cite{shapenet}. Despite being pretrained on a less diverse set of objects and having the smallest size, DuoMamba demonstrates competitive performance. Among models with fewer than 50M parameters (DuoMamba, PointBERT \cite{pointbert}, SparseConv \cite{sparseconv}), our model outperforms all others by a significant margin of 3\% in zero-shot accuracy. While Uni3D-giant \cite{uni3d} achieves a slightly higher accuracy with a gap of 1.8\%, it comes at the cost of a substantially larger model size, with 1016.5M parameters -- 35 times the size of DuoMamba. This highlights the optimal balance between model size and performance offered by our method compared to existing approaches.}    
    \label{fig:supp_zeroshot_comparison}
\end{figure}

\paragraph{Model Size and Zero-Shot Object Classification Performance.} We compare the parameter counts of various point cloud encoders and their zero-shot performance in Figure \ref{fig:supp_zeroshot_comparison}. Despite only being pretrained on ShapeNetCore \cite{shapenet}, our DuoMamba outperforms all existing models of comparable size that are pretrained on 880K 3D objects -- 17 times more data. Notably, the zero-shot accuracy gap between our model and the best-performing model Uni3D-giant \cite{uni3d} is just 1.8\%, even though our model is only 1/35 its size. This highlights DuoMamba's superior size-to-performance efficiency. Scaling up the model and pretraining on larger datasets is likely to further enhance performance, which we leave as future work.

\begin{figure}[!h]
    \centering
    \includegraphics[width=0.8\linewidth]{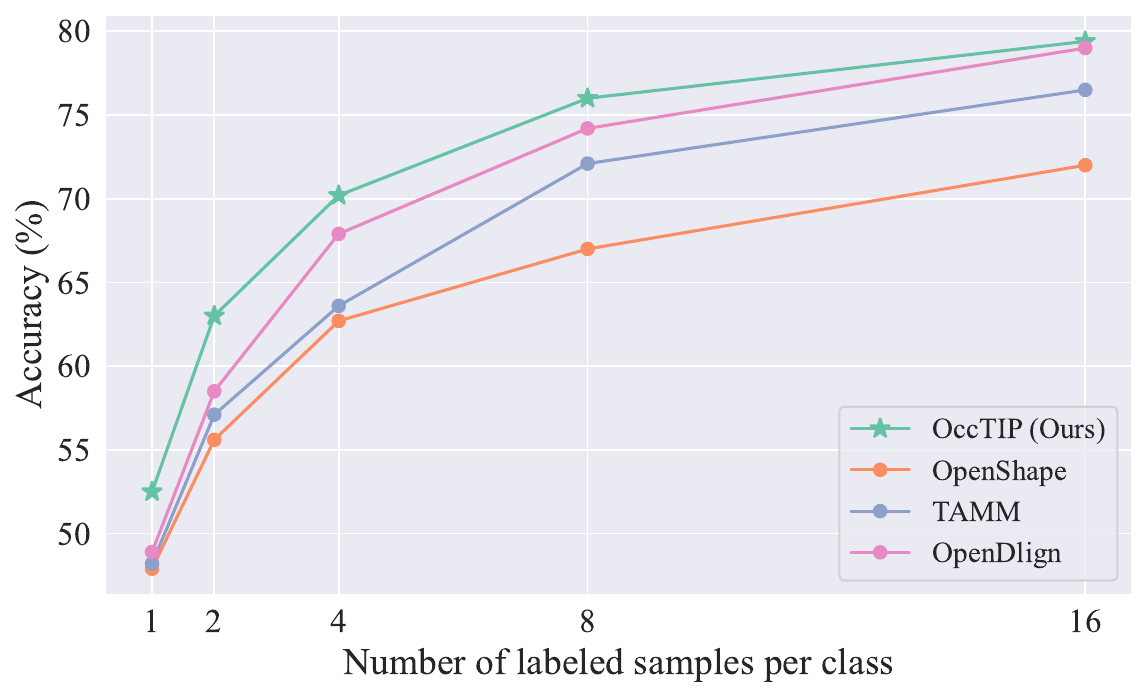}
    \caption{\textbf{Few-shot linear probing on ScanObjectNN} \cite{scanobjectnn}. Our method is pretrained on 52K ShapeNetCore \cite{shapenet} objects, whereas other models are pretrained on 880K objects. Despite using significantly less data, our framework OccTIP outperforms all existing methods across all few-shot settings, demonstrating the data efficiency and the high-quality latent space learned by our approach.
    }    
    \label{fig:supp_fewshot_comparison}
\vspace{-1mm}
\end{figure}

\vspace{-5mm}
\paragraph{Few-Shot Linear Probing.} We perform a few-shot experiment similar to the one in Section \ref{main_exp_fewshot} 
(main paper), this time comparing our approach against models pretrained on the ensemble of 880K 3D objects. As illustrated in Figure \ref{fig:supp_fewshot_comparison}, our method consistently outperforms all other works across all few-shot settings, highlighting our pretraining framework's data efficiency and effectiveness in learning robust and generalizable features for real-world recognition.

\vspace{-1cm}
\section{Additional Quantitative Results}
\begin{figure}[!h]
    \centering
    \includegraphics[width=0.8\linewidth]{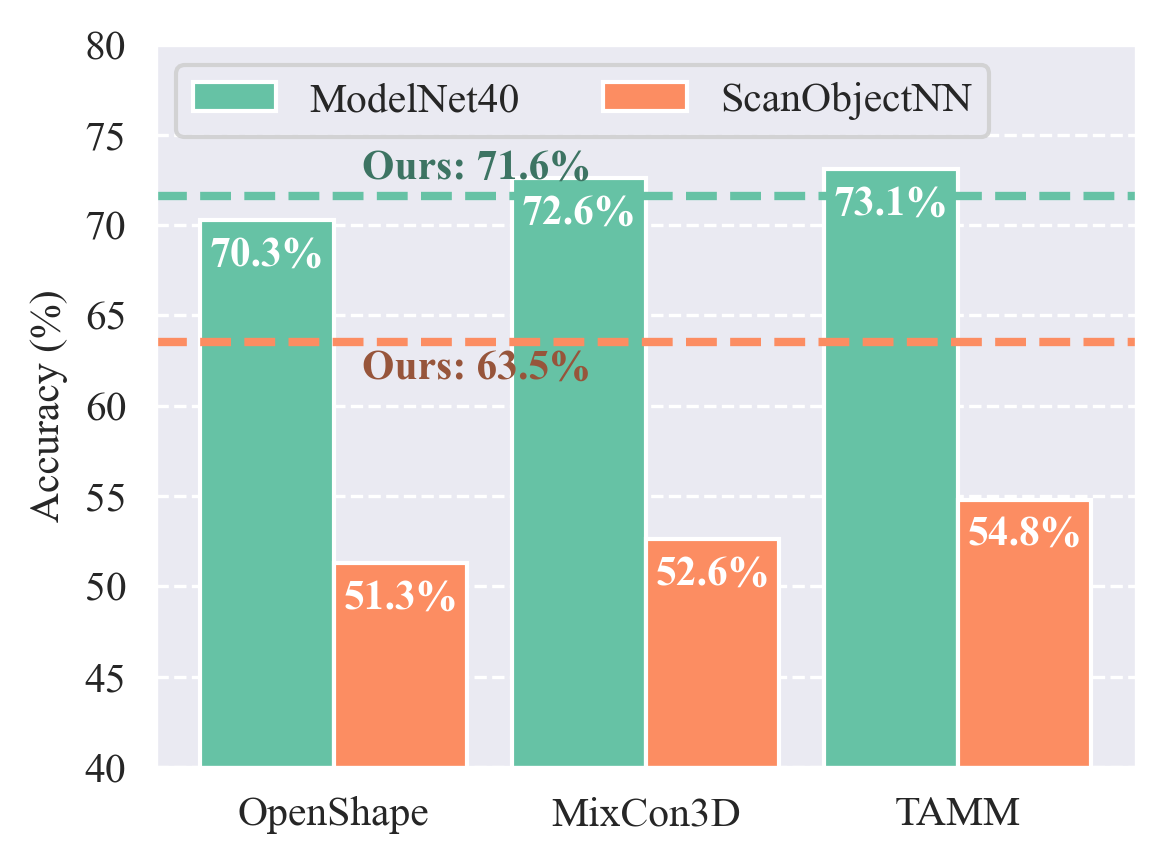}
    \caption{Comparison with methods pretrained on \textit{complete} point clouds.}    
    \label{fig:duomamba_modelnet40}
\vspace{-1mm}
\end{figure}

\paragraph{Evaluate Pretrained DuoMamba on ModelNet40.} To evaluate {DuoMamba (pretrained with OccTIP) on complete point clouds,} we generate partial point clouds from 12 views (as in pretraining) and use majority voting for class prediction. Figure \ref{fig:duomamba_modelnet40} shows that on ModelNet40, we perform competitively with previous works pretrained on full point clouds and even \textbf{surpass OpenShape} by 1.3\%. 

\paragraph{Complete Results for Zero-Shot 3D Object Detection.}
\label{sec:supp_complete_obj_det}
The average precision (AP) for each class and the mean Average Precision (mAP) for the zero-shot 3D object detection experiments (Section \ref{subsec:obj_det} in the main paper) are provided in Table \ref{tab:supp_obj_det_scannetv2} (for ScanNetV2 \cite{scannetv2} benchmark) and Table \ref{tab:supp_obj_det_sunrgbd} (for SUN RGB-D \cite{sunrgbd} benchmark). 
Our method OccTIP consistently achieves the best or second-best AP across most categories and achieves the highest mAP, with a significant margin over existing techniques on both datasets. These results highlight the effectiveness of OccTIP and its applicability to complex, real-world recognition tasks.

\paragraph{Pretraining with Complete vs. Partial Point Clouds.} Table \ref{tab:rebuttal_ablation_full_partial} shows that our synthetic partial data consistently improves all models' accuracy on real-world ScanObjectNN, with DuoMamba performing best in both settings.
\begin{table}[h!]
\centering
\setlength\aboverulesep{0pt}\setlength\belowrulesep{0pt}
    \setlength{\tabcolsep}{6.5pt}  
    \resizebox{0.4\textwidth}{!}{
    \begin{tabular}{c|cc|c}
        \toprule
        Pretraining data & SparseConv & PointBERT & \cellcolor{gray!20}DuoMamba \\
        \midrule
        Complete & 56.0 & 55.5 & \cellcolor{gray!20}\textbf{57.5}  \\
        \rowcolor{gray!20}
        Partial (OccTIP) & 61.7 (+5.7) & 60.6 (+5.1) & \textbf{63.5 (+6.0)} \\ 
        \bottomrule
    \end{tabular}
    }
\caption{ScanObjectNN accuracy when pretraining with full vs partial data.}
\label{tab:rebuttal_ablation_full_partial}
\end{table}

\paragraph{Architecture Influence on Object Detection Performance.} 
Table \ref{tab:rebuttal_obj_det} compares object detection performance of DuoMamba and PointBERT pretrained with OccTIP against PointBERT’s best performance by previous pretraining baselines. OccTIP consistently enhances PointBERT’s performance, and its combination with DuoMamba achieves the best results.
\begin{table}[h!]
\centering
\setlength\aboverulesep{0pt}\setlength\belowrulesep{0pt}
    \setlength{\tabcolsep}{6.5pt}  
    \resizebox{0.4\textwidth}{!}{
        \begin{tabular}{c|c|cc|cc}
        \toprule
        Pretraining & \multirow{2}{*}{Encoder} & \multicolumn{2}{c|}{ScanNetV2} & \multicolumn{2}{c}{SUN RGB-D} \\ \cline{3-6} 
        method  &          & mAP$_{25}$ & mAP$_{50}$ & mAP$_{25}$ &  mAP$_{50}$ \\ 
        \midrule
        Best current & PointBERT & 24.1 & 19.1 & 18.9 & 10.0 \\ 
        \rowcolor{gray!20}
        OccTIP &  &  25.4 & 19.3 & 21.9 & 11.7 \\
        \midrule
        \rowcolor{gray!20}
        OccTIP & DuoMamba & \textbf{28.9} & \textbf{22.7} & \textbf{24.4} & \textbf{13.0} \\ 
        \bottomrule
        \end{tabular}
    }
\caption{Detection results of different models and pretraining methods.}
\label{tab:rebuttal_obj_det}    
\end{table} 

\end{document}